\documentclass[10pt,twocolumn,letterpaper]{article}

\usepackage{cvpr}
\usepackage{times}
\usepackage{epsfig}
\usepackage{graphicx}
\usepackage{amsmath}
\usepackage{comment}
\usepackage{amssymb}

\def\manifold{QATM }

\newcommand{\img}[1]{$\mathbf{#1}$}
\usepackage{subcaption}
\usepackage{listings}
\usepackage{color}
\usepackage{xcolor}
\usepackage{algorithm}
\usepackage[noend]{algpseudocode}
\usepackage[pagebackref=true,breaklinks=true,colorlinks,bookmarks=false]{hyperref}

\definecolor{codegreen}{rgb}{0,0.6,0}
\definecolor{codegray}{rgb}{0.5,0.5,0.5}
\definecolor{codepurple}{rgb}{0.58,0,0.82}
\definecolor{backcolour}{rgb}{0.95,0.95,0.92}

\lstdefinestyle{mystyle}{
  backgroundcolor=\color{backcolour},   commentstyle=\color{codegreen},
  keywordstyle=\color{magenta},
  numberstyle=\tiny\color{codegray},
  stringstyle=\color{codepurple},
  basicstyle=\footnotesize,
  breakatwhitespace=false,         
  breaklines=true,                 
  captionpos=b,                    
  keepspaces=true,                 
  numbers=left,                    
  numbersep=5pt,                  
  showspaces=false,                
  showstringspaces=false,
  showtabs=false,                  
  tabsize=2
}

\cvprfinalcopy % *** Uncomment this line for the final submission
\def\cvprPaperID{6347} % *** Enter the CVPR Paper ID here

\ifcvprfinal\fi
\begin{document}

%%%%%%%%% TITLE
\title{QATM: Quality-Aware Template Matching For Deep Learning}

\author{
Jiaxin Cheng \quad Yue Wu \quad Wael Abd-Almageed \quad Premkumar Natarajan\\
USC Information Sciences Institute, Marina del Rey, CA, USA\\
{\tt\small chengjia@\{usc/isi\}.edu \{yue\_wu,wamageed,pnataraj\}@isi.edu} 
}


\maketitle
%\thispagestyle{empty}

%%%%%%%%% ABSTRACT
\begin{abstract}
   Finding a template in a search image is one of the core problems  many computer vision,  such as semantic image semantic, image-to-GPS verification \etc. We propose a novel quality-aware template matching method, QATM,  which is not only used as a standalone template matching algorithm, but also a trainable layer that can be easily embedded into any deep neural network. Here, our quality can be interpreted as the distinctiveness of matching pairs. Specifically, we assess the quality of a matching pair using soft-ranking among all matching pairs, and thus different matching scenarios such as 1-to-1, 1-to-many, and many-to-many will be all reflected to different values. Our extensive evaluation on classic template matching benchmarks and deep learning tasks demonstrate the effectiveness of QATM. It not only outperforms state-of-the-art template matching methods when used alone, but also largely improves existing deep network solutions. 
\end{abstract}

%%%%%%%% Introduction
\section{Introduction and Review}
Template matching is one of the most frequently used techniques in computer vision applications, such as video tracking\cite{zhang2014partial, zhang2015robust, adam2006robust, comaniciu2003kernel}, image mosaicing \cite{szeliski2007image, brown2007automatic}, object detection \cite{felzenszwalb2010object, coughlan2000efficient, yuille1992feature}, character recognition \cite{trier1996feature, alaei2014complete, ryan2015examination, boia2016logo}, and 3D reconstruction \cite{scharstein2002taxonomy, seitz2006comparison, malti2013monocular}. Classic template matching methods often use sum-of-squared-differences (SSD) or normalized cross correlation (NCC) to calculate a similarity score between the template and the underlying image. These approaches work well when the transformation between the template and the target search image is simple. However, these methods start to fail when the transformation is complex or non-rigid, which is common in real-life. In addition, other factors, such as occlusions and color shifts, make these methods even more fragile. 

Numerous approaches have been proposed to overcome these real-life difficulties applying standard template matching. 
Dekel et al. \cite{dekel2015best} introduced the Best-Buddies-Similarity (BBS) measure, which focuses on the nearest-neighbor (NN) matches to exclude potential and bad matches caused by the background pixels. Deformable Diversity Similarity (DDIS) was introduced in \cite{talmi2017template}, which explicitly considers  possible template deformation and uses the diversity of NN feature matches between a template and a potential matching region in the search image. Co-occurrence based template matching (CoTM) was introduced in \cite{kat2018matching} to quantify the dissimilarity between a template and a potential matched region in the search image. These methods indeed improve the performance of template matching. However, these methods cannot be used in deep neural networks (DNN) because of two limitations --- (1) using non-differentiable operations, such as thresholding, counting, \etc. and (2) using operations that are not efficient with DNNs, such as loops and other non-batch operations. 

%In contrast, more and more compute vision problems are now no longer solved by a heuristic algorithm but a deep neural network, since the great success of AlexNet~\cite{krizhevsky2012imagenet} in 2012. This means unless a template matching method is DNN compatible, it will no longer be used as a part of the solution.

Existing DNN-based methods use simple methods to mimic the functionality of template matching~\cite{luo2016efficient,wu2017deep,thewlis2016fully,tang2016multi,bai2016exploiting}, such as computing the tensor dot-product~\cite{rocco2017convolutional}\footnote{See \texttt{numpy.tensordot} and \texttt{tensorflow.tensordot}.} between two batch tensors of sizes $B\times H\times W\times L$ and $B\times H'\times W'\times L$ along the feature dimension (\ie, $L$ here), and producing a batch tensor of size $B\times H\times W\times H'\times W'$ containing all pairwise feature dot-product results. Of course, additional operations like max-pooling may also be applied~\cite{wu2017deep,BusterNet,rocco2017convolutional,cheng2018ACCV}.

%These efforts clearly achieve great success with the use only coarse template matching techniques. However, isn't this another reason urging us to propose a new template matching layer to further improve various existing DNN solutions? 

In this paper, we propose the quality-aware template matching (QATM) method, which can be used as a standalone template matching algorithm, or in a deep neural network as a trainable layer with learnable parameters. It takes the uniqueness of pairs into consideration rather than simply evaluating matching score. \manifold is composed of differentiable and batch-friendly operations and, therefore, is efficient during DNN training. More importantly, \manifold is inspired by assessing the matching quality of source and target templates, and thus is able handle different matching scenarios including 1-to-1, 1-to-many, many-to-many and no-matching. Among different matching cases, only the 1-to-1 matching is considered to be high quality due to it's more distinctive than 1-to-many and many-to-many cases.

The remainder of paper is organized as follows. Section ~\ref{sec:qatm} discusses motivations and introduces \manifold. In Section \ref{sec:exptTP}, the performance of \manifold is studied in classic template matching setting. \manifold is evaluated on both semantic image alignment and image-to-GPS verification problems in Section~\ref{sec:qtam-deep}. We conclude the paper and discuss future works in Section ~\ref{sec:conclusion}.

%For example, the constrained image splicing detection problem~\cite{wu2017deep} is a typical implicit template matching problem --- given a query image \img{Q} and a potential donor image \img{P}, find the spliced region in \img{P} that belongs to \img{Q}. The image-to-GPS problem is another example,

%\input{latex/introduction.tex}

%%%%%%%%% METHOD
\section{Quality-Aware Template Matching}\label{sec:qatm}
\subsection{Motivation}
In computer vision, regardless of the application, many methods implicitly attempt the solve some variant of following problem ---  \textit{given an exemplar image (or image patch), find the most similar region(s) of interest in a target image}. Classic template matching \cite{dekel2015best, talmi2017template, kat2018matching}, constrained template matching \cite{BusterNet}, image-to-GPS matching \cite{cheng2018ACCV}, and semantic alignment \cite{rocco2017convolutional, rocco2018end, choy2016universal, Han_2017_ICCV} methods all include some sort of template matching, despite differences in the details of each algorithm. Without loss of generality, we will focus the discussion on the fundamental template matching problem, and illustrate applicability to different problem in later sections.

One known issue in most of existing template matching methods is that typically, all pixels (or features) within the
template and a candidate window in the target image are
taken into account when measuring their similarity\cite{dekel2015best}. This is undesirable in many cases, for example when the background behind the object of interest changes between the
template and the target image. To overcome this issue, the BBS \cite{dekel2015best} method relies on  nearest neighbor (NN) matches between the template and the target, so that it could exclude most of  background pixels for matching. On top of BBS, the DDIS \cite{talmi2017template} method uses the additional deformation information in NN field, to further improve the matching performance. 

Unlike previous efforts, we consider five different template matching scenarios, as shown in Table~\ref{tab:cases}, where $t$ and $s$ are patches in the template \img{T} and search \img{S} images, respectively. Specifically, ``1-to-1 matching'' indicates exact matching, i.e. two matched objects,  ``1-to-$N$'' and ``$M$-to-1'' indicates $s$ or $t$ is a homogeneous or patterned patch causing multiple matches, \eg a sky or a carpet patch, and ``$M$-to-$N$'' indicates many homogeneous/patterned patches both in \img{S} and \img{T}. It is important to note that  this formulation is completely different from the previous NN based formulation, because even though $t$ and $s$ are nearest neighbors, their actual relationship still can be any of the five cases considered. Among four matching cases, only 1-to-1 matching is considered as high quality. This is due to the fact that in other three matching cases, even though pairs may be highly similar, that matching is less distinctive because of multiple matched candidates. Which turned out lowering the reliability of that pair.

\begin{table}[!h]
    \centering
    \small
    \setlength\tabcolsep{2pt}
    \begin{tabular}{c|cccc|c}
        \hline
         &\multicolumn{4}{c|}{\textbf{Matching Cases}} & \textbf{Not}\\  \cline{2-5}
          &  1-to-1 & 1-to-$N$ & $M$-to-1 & $M$-to-$N$ &  \textbf{Matching} \\ \hline
         Quality & High & Medium & Medium & Low & Very Low \\ \hline 
         QATM $(s,t)$ & 1 & 1/$N$ & $1/M$ & $1/MN$ & $1/\|T\|\|S\|\approx 0$\\ \hline
    \end{tabular}
    \caption{Template matching cases and ideal scores.}
    \label{tab:cases}
\end{table}  

It is clear that the ``1-to-1'' matching case is the most important, while the ``not-matching'' is almost useless. It is therefore not difficult to come up the qualitative assessment for each case in the Table~\ref{tab:cases}. As a result, the optimal matched region in \img{S} can be found as the place that maximizes the overall matching quality. We can therefore come up with a quantitative assessment of the matching as shown in Eq.~\eqref{eq:roi}
\begin{equation}\label{eq:roi}
    R^* =  \mathop{\arg\max}_{R}\{ \sum_{r\in R} \max\big\{\textrm{Quality}(r,t) | t\in \mathbf{T}\big\} \}
\end{equation}
such that the region $R$ in \img{S} that maximizes the overall matching quality will be the optimally matching region. $R$ is a fixed size candidate window and we used the size of object as window size in the experiment.

%===========================================================================
\subsection{Methodology}
%===========================================================================
To make Eq.~\eqref{eq:roi} applicable to template matching, we need to define Quality($s,t$), \ie how to assess the matching quality between $(s,t)$. In the rest of section, we derive the quality-aware template matching (QATM) measure, which is a proxy function of the ideal quality assessment Quality($s,t$). 

Let $f_s$ and $f_t$ be the feature representation of patch $s$ and $t$, and $\rho(\cdot)$ is a predefined similarity measure between two patches, \eg cosine similarity. Given a search patch $s$, we define the likelihood function that a template patch $t$ is matched, as shown in Eq.~\ref{eq:likelihood},
\begin{equation}\label{eq:likelihood}
    L(t|s) = \frac{\exp\{\alpha\cdot\rho(f_t,f_s)\}}{\sum_{t'\in\mathbf{T}}\exp\{\alpha\cdot\rho(f_{t'},f_s)\}}
\end{equation}
where $\alpha$ is a positive number and will be discussed later. This likelihood function can be interpreted as a soft-ranking of the current patch $t$ compared to all other patches in the template image in terms of matching quality. It can be alternatively considered as a heated-up softmax embedding~\cite{zhang2018heated}, which is the softmax activation layer with a learnable temperature parameter, \ie $\alpha$ in our context.

In this way, we can define the QATM measure as simple as the product of likelihoods that $s$ is matched in \img{T} and $t$ is matched in \img{S} as shown in Eq.~\eqref{eq:qatm}.
\begin{equation}\label{eq:qatm}
    \textrm{QATM}(s,t) = L(t|s) \cdot L(s|t)
\end{equation}

\begin{table}[!h]
    \centering
    \small
    \begin{tabular}{c|cc|c}\hline
     \textbf{Matching Case} & $L(s|t)$ & $L(t|s)$ &  QATM$(s,t)$\\\hline
        1-to-1 &  1 & 1 & 1\\
        1-to-$N$ & 1 & 1/$N$ & 1/$N$ \\
        $M$-to-1 & 1/$M$ & 1 & 1/$M$ \\
        $M$-to-$N$ & 1/$M$ & 1/$N$ & 1/$MN$\\
        Not Matching & 1/$\|\mathbf{S}\|$ & 1/$\|\mathbf{T}\|$ & $\approx 0$ \\\hline
    \end{tabular}
    \caption{Ideal QATM scores}
    \label{tab:qatm}
\end{table}

Any reasonable similarity measure $\rho(\cdot)$ that gives a high value when $f_t$ and $f_s$ are similar, a low value otherwise could be used. When $t$ and $s$ truly matched, $\rho(f_t,f_s)$ should be larger than those unmatched cases $\rho(f_t,f_{s'})$. Equivalently, this means $\rho(f_t,f_s)$ is the best match and thus the maximum score. This score will ideally be 1, after lifting by $\alpha$ and activating by the softmax function, when appropriate $\alpha$ parameter is selected. Similarly, when $t$ matches $N$ of $s$ patches, we should have $N$ equally high matching scores, indicating $L(s|t)=1/N$ in the ideal case. Table~\ref{tab:qatm} summarizes the ideal scores of all five cases, and their values match the subjective quality assessment on individual cases shown in Table~\ref{tab:cases}. Once we have the pairwise QATM results between \img{S} and \img{T}, the matching quality of an ROI $s$ can be found as shown in Eq.~\eqref{eq:qs}
\begin{equation}\label{eq:qs}
    q(s) = \max\big\{\textrm{QATM}(s,t)| t\in \mathbf{T}\big\}
\end{equation}
where $q(\cdot)$ indicates the matching quality function. Eventually, we can find the best matched region $R^*$ which maximizes the overall matching quality as shown in Eq.~\eqref{eq:qatmroi}.
\begin{equation}\label{eq:qatmroi}
    R^* =  \mathop{\arg\max}_{R}\left\{ \sum_{r\in R} q(r)\right\}%=\\ \mathop{\arg\max}_{R\in\mathbf{S}}\left\{ \sum_{r\in R} \max\left\big\{\textrm{QATM}(r,t)| t\in \mathbf{T}\right\big\} \right\}
\end{equation}

%===========================================================================
\subsection{QATM As An Algorithmic DNN Layer}
%===========================================================================
Proposed QATM assesses the matching quality in a continuous way. Therefore, its gradients can be easily computed via the chain rule of individual function (all of which can be implemented through either a standard DNN layer \eg \texttt{softmax} activation, or basic mathematical operators provided in most of DNN frameworks).
\begin{algorithm}
\caption{Compute QATM and matching quality between two images}\label{alg:qatm}
\begin{algorithmic}[1]
\State \textbf{Given:} template image $I_T$ and search image $I_S$, a feature extracting model $F$, a temperature parameter $\alpha$. Func($\cdot|I$) indicates doing operation along axis of $I$.
\State $T\gets F(I_T)$
\State $S\gets F(I_S)$
\State $\rho_{st}\gets$ Patch-wiseSimilarity($T, S$)\Comment{Which can be easily obtained by off-the-shelf functions such as \texttt{tensorflow.einsum} or \texttt{tensorflow.tensordot}}
\State $\rho_{st}\gets \rho_{st} \times \alpha$
\State $L(s|t)\gets$ Softmax($\rho_{st}|T$)
\State $L(t|s)\gets$ Softmax($\rho_{st}|S$)
\State $QATM \gets L(s|t) \times L(t|s)$
\State $S_{map}\gets$ Max($QATM|T$) \Comment{Matching quality score}
\State $T_{map}\gets$ Max($QATM|S$)
\end{algorithmic}
\end{algorithm}

In Alg.~\ref{alg:qatm}, we demonstrate how to compute the matching quality map form both \img{T} and \img{S}. One can easily implement it into DNN in roughly 30 lines of \texttt{Python} code using deep learning librarys such as \texttt{Tensorflow} and \texttt{Pytorch}. Specifically, we use the \textit{cosine similarity} as an example to assess the raw patch-wise similarity, \texttt{tf.einsum}(line 4) computes all patch-wise similarity scores in a batch way. Once QATM$(t,s)$ is computed, we can compute the template matching map for the template image \img{T} and the target search image \img{S}, respectively, as shown in lines 9 --- 10.  As one can see, when the $\alpha$ parameter is not trainable, \ie a fixed value, then the proposed QATM layer degrades to a classic template matching algorithm. 

%===========================================================================
\subsection{Discussions on $\alpha$}~\label{sec:alpha}
%===========================================================================
 In this section, we discuss how $\alpha$ should be picked in a direct template matching scenario that does not involve training a DNN. We later show that QTAM can easily be embedded as a trainable layer in DNNs to perform template matching without manual tuning structure according to tasks. 
 
When applying Eq.~\eqref{eq:likelihood}, $\alpha$ serves two purposes --- (1) matched patches will have ranking scores as close to 1 as possible, and (2)  unmatched patches will have ranking scores as close to 0 as possible. As one can see, as $\alpha$ increases, $L(t|s)^+$, the likelihood of matched cases, will also increase, and will quickly reach its maximum of 1 after some $\alpha$. However, this does not mean we can easily pick a large enough $\alpha$, because a very large $\alpha$ will also push $L(t|s)^-$, the likelihood of unmatched cases, to deviate from 0. Therefore, a good $\alpha$ choice can be picked as the one that provides the largest quality discernibility as shown in Eq.~\eqref{eq:bestAlpha}
\begin{equation}\label{eq:bestAlpha}
    \alpha^*=\mathop{\arg\max}_{\alpha>0}\left\{L(t|s)^+-L(t|s)^-\right\}.
\end{equation}
In practice, it is difficult to manually set $\alpha$ properly without knowing details about the similarity score distributions of both matched and unmatched pairs. If both distributions are known, however, we can simulate both $L(t|s)^+$ and $L(t|s)^-$. Without loss of generality, say there are $N$ patches in \img{T}. $L(t|s)$, whether or not $(t,s)$ is the matched pair, can be obtained by simulating one $f_t$ feature and $N$ of $f_s$ feature, or equivalently, by simulating $N$ number of $\rho(f_t,f_s)$ similarity scores according to its definition Eq.~\eqref{eq:likelihood}. The major difference between the matched and unmatched cases is that we need one score from the score distribution of matched pairs and $N-1$ scores from the distribution of unmatched pairs for $L(t|s)^+$, while all $N$ scores from the distribution of unmatched pairs for $L(t|s)^-$. 

\begin{figure}
    \centering
    \includegraphics[width=.6\linewidth]{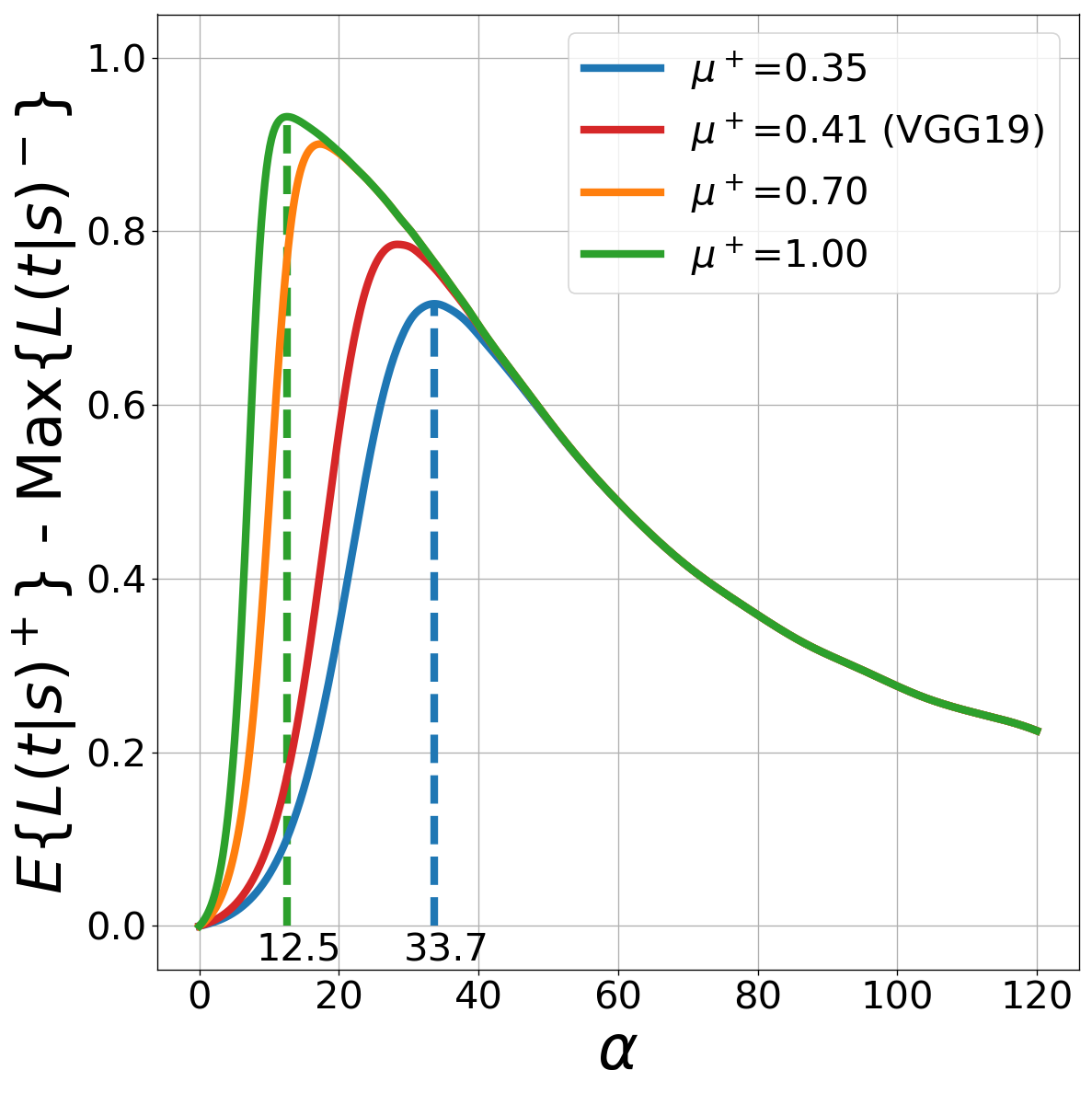}
    \caption{The quality discernibility for varying $\alpha$.}
    \label{fig:diff}
\end{figure}

Fig.~\ref{fig:diff} shows the difference between $\mathbb{E}[L(t|s)^+]$ and $\max\{L(t|s)^-\}$ for different $\alpha$ values, when the genuine and imposter scores follow the normal distribution $\mathcal{N}(\mu^+,0.01)$ and $\mathcal{N}(0,0.05)$ for $N=2200$. As one can see, the difference plot is uni-modal, and the optimal $\alpha$ increases as the mean $\mu^+$ decreases. 
This figure is more meaningful when the used feature is from a DNN and the used raw similarity measure is the cosine similarity. Zhang et al. \cite{zhang2017learning} provides the theoretical cosine similarity score distribution for unmatched pairs, whose mean is 0 and variance is $1/d$, where $d$ is the feature dimension. Our empirical studies shows that many DNN features attains $\mu^+$ above 0.3, \eg the VGG19 feature. Consequently, a reasonable $\alpha$ for DNN features is roughly in $[12.5, 33.7]$ when cosine similarity is used. 

%===========================================================================
%===========================================================================

%%%%%%%% EXPT
\section{QATM Performance in Template Matching}\label{sec:exptTP}
We start with evaluating the proposed QATM performance on the classic template matching problem. Our code is released in the open repository \url{https://github.com/cplusx/QATM}.

\subsection{Experimental Setup}
To find the matched region in the search image \img{S}, we compute the matching quality map on \img{S} through the proposed \texttt{NeuralNetQATM} layer (without learning $\alpha$) (see Alg.~\ref{alg:qatm}), which takes a search image $I_\mathbf{S}$ and a template image $I_\mathbf{T}$ as inputs. One can therefore find the best matched region $R^*$ in \img{S} using Eq.~\eqref{eq:qatmroi}. 

We  follow the evaluation process given in ~\cite{simakov2008summarizing} and use the standard OTB template matching dataset~\cite{wu2013online}, which contains 105 template-image pairs from 35 color videos. We use the 320-d convoluational feature from a pretrained ImageNet-VGG19 network. The standard intersection over union (IoU) and the area-under-curve (AUC) methods are used as evaluation metrics. QTAM is compared against three state-of-the-art methods, BBS~\cite{dekel2015best}, DDIS~\cite{talmi2017template} and CoTM~\cite{simakov2008summarizing}, plus the classic template matching using SSD and NCC.

\subsection{Performance On The Standard OTB Dataset}
\begin{figure*}[!h]
    \centering\small
    \begin{tabular}{cc|c}
         \includegraphics[width=.30\linewidth]{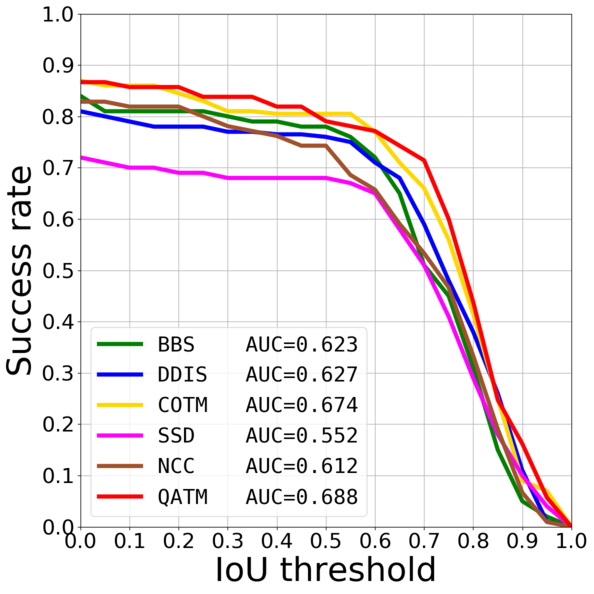}
         &
         \includegraphics[width=.31\linewidth]{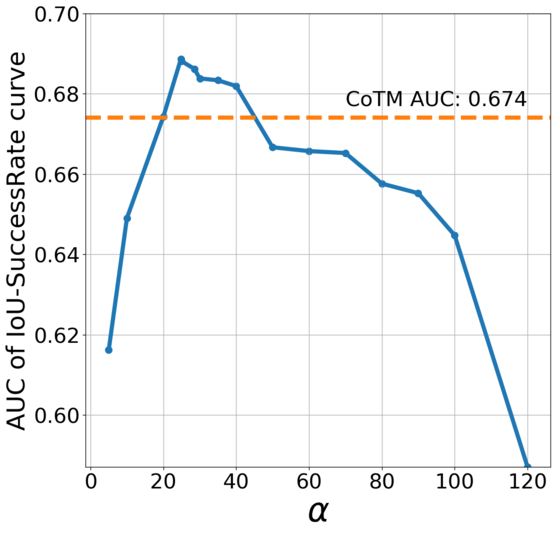}
         &
         \includegraphics[width=.31\linewidth]{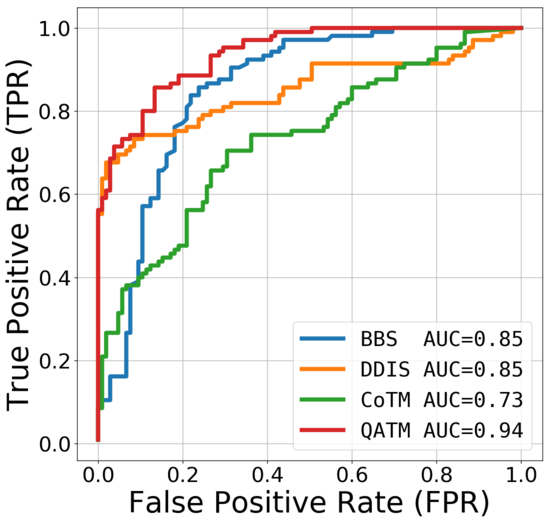}\\
         (a) & (b) & (c)\\
    \end{tabular}
    \caption{Template matching performance comparisons. (a) QATM v.s. SOTA methods on the OTB dataset. (b) QATM performance under varying $\alpha$ on the OTB dataset. (c) QATM v.s. SOTA methods on the MOTB dataset.}
    \label{fig:sotaTemplate}
\end{figure*}

In this experiment, we  follows all the experiment settings from ~\cite{kat2018matching}, and evaluates the proposed QATM method on the standard OTB dataset. The $\alpha$ value is set to 28.4, which is the peak of {\it VGG}'s curve (see Fig.~\ref{fig:diff}).
The QATM performance as well as all baseline method performance are shown in Fig.~\ref{fig:sotaTemplate}-(a). As one can see, the proposed QATM outperforms state-of-the-art methods and lead the second best (CoTM) by roughly 2\% in terms of AUC score, which is clearly a noticeable improvement when comparing to the 1\% performance gap between BBS and its successor DDIS. 

Since the proposed QATM method has the parameter $\alpha$, we evaluate the QATM performance under varying $\alpha$ values as shown in Fig.~\ref{fig:sotaTemplate}-(b). It is clear that the overall QATM performance is not very sensitive to the choice of value when $\alpha$ is around optimal solution. As indicated by the horizontal dash line in Fig.~\ref{fig:sotaTemplate}-(b), a range of $\alpha$ (rather than a single value) leads to better performance than the state-of-the-art methods. More qualitative results can be found in Fig.~\ref{fig:motbResults}.

\subsection{Performance On The Modified OTB Dataset}
One issue in the standard OTB dataset is that it does not contain any negative samples, but we have no idea whether a template of interest exist in a search image in real-applications. We therefore create a modified OTB (MOTB) dataset. Specifically, for each pair search image \img{S} and template \img{T} in OTB, we (1) reuse this pair (\img{S},\img{T}) in MOTB as a positive sample and (2) keep \img{S} untouched while replacing \img{T} with a new template \img{T'}, where \img{T'} is from a different OTB video, and use this (\img{S},\img{T'}) as a negative sample. The negative template \img{T'} is chosen to be the same size as \img{T} and is randomly cropped from a video frame.

The overall goal of this study is to fairly evaluate the template matching performance with the presence of negative samples. For each sample in MOTB, a pair of (template, search image), we feed it to a template matching algorithm and record the average response of the \textit{found} region in a search image. For the proposed QATM method, we again use $\alpha=28.4$.  These responses along with the true labels of each pairs are then used to plot the AUC curves shown in Fig.~\ref{fig:sotaTemplate}-(c). Intuitively, a good template matching method should give much lower matching scores for a negative sample than for a positive sample, and thus attain a higher AUC score. The proposed QATM method obviously outperform the three state-of-the-art methods by a large margin, which is roughly 9\% in terms of AUC score. More importantly, the proposed QATM method clearly attains much higher true positive rate at low false positive rates. This result is not surprising since the proposed QATM is quality aware. For example, when a negative template is homogeneous, all methods will find a homogeneous region in the search image since it is the most similar region. The difference is that our approach is quality-aware and thus the matching score of this type will be much lower than that of a positive template, while other methods do not have this feature.

%===================================================================================
\subsection{Discussions}
\begin{figure}[!h]
    \centering
    \includegraphics[width=.995\linewidth]{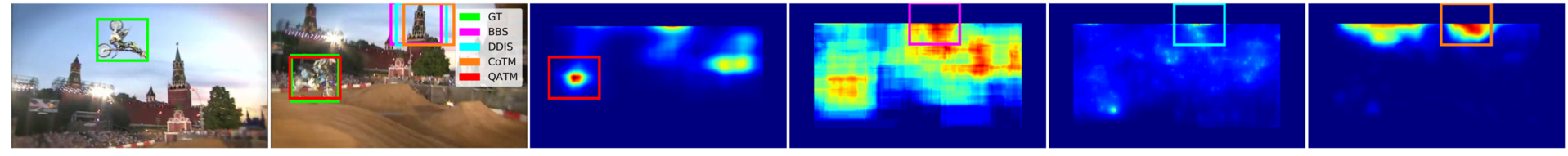}\\
    \includegraphics[width=.995\linewidth]{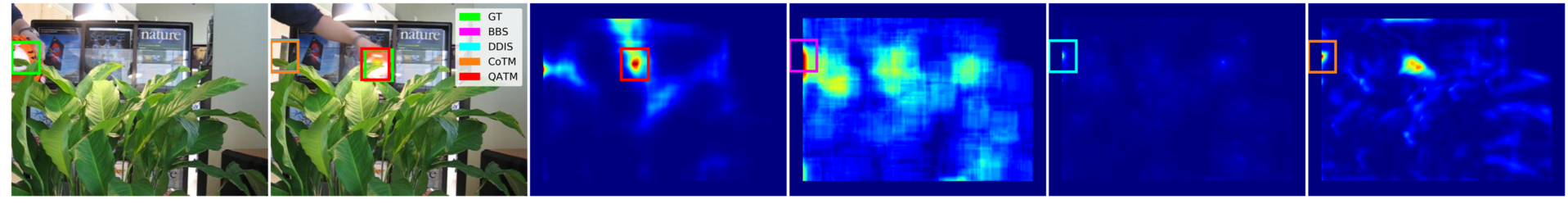}\\
    \includegraphics[width=.995\linewidth]{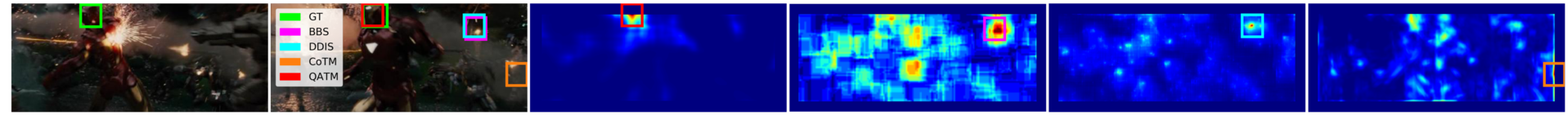}\\
    \includegraphics[width=.995\linewidth]{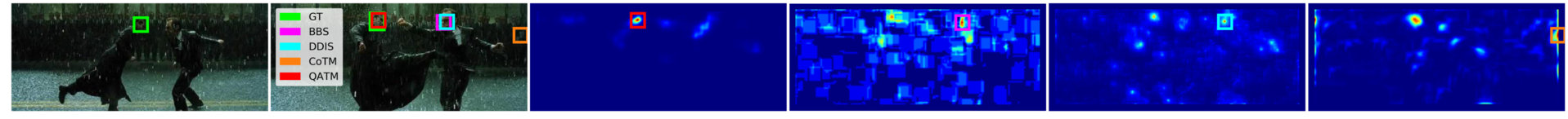}\\
%    \,\\
%    \hline\\
%    \,\\
    \includegraphics[width=.995\linewidth]{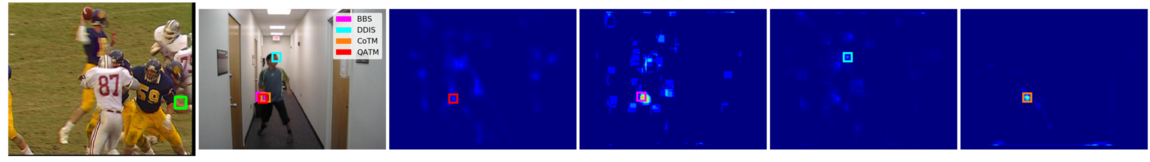} \\
    \includegraphics[width=.995\linewidth]{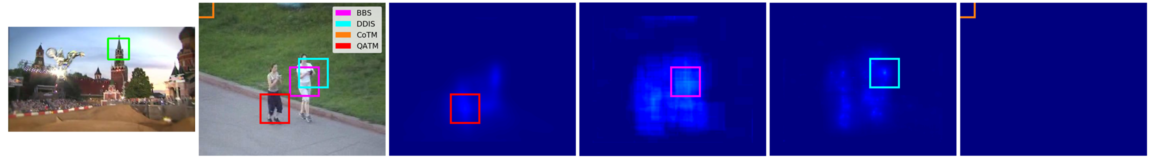} \\
    \includegraphics[width=.995\linewidth]{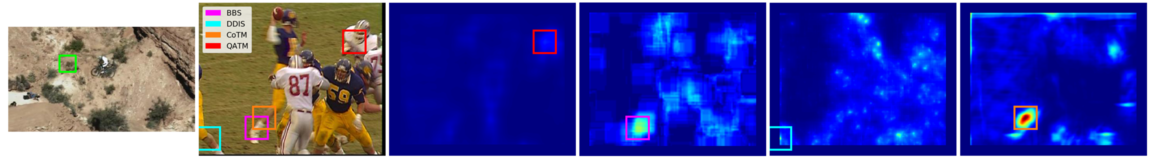} \\
    \includegraphics[width=.995\linewidth]{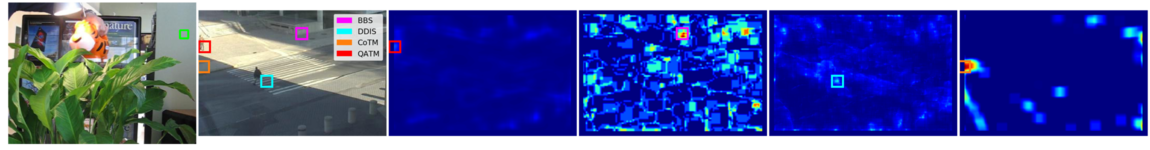} \\
    \begin{tabular*}{\linewidth}{c @{\extracolsep{\fill}} ccccc}
        \small{\textcolor{green}{Template}}&
        \small{Search} &
        \small{\textcolor{red}{QATM}} & 
        \small{\textcolor{magenta}{BBS}}&
        \small{\textcolor{cyan}{DDIS}}  &
        \small{\textcolor{orange}{CoTM}} 
    \end{tabular*}
\caption{Qualitative template matching performance. Columns from left to right are: the \textcolor{green}{template} frame, the target search frame with predicted bounding boxes overlaid (different colors indicate different method), and the response maps of \textcolor{red}{QATM}, \textcolor{magenta}{BBS}, \textcolor{cyan}{DDIS}, \textcolor{orange}{CoTM}, respectively. Rows from top to bottom: the top four are positive samples from OTB, while the bottom four are negative samples from MOTB. Best viewed in color and zoom-in mode. }
    \label{fig:motbResults}
\end{figure}

%===================================================================================
Fig.~\ref{fig:motbResults} provides more qualitative results from the proposed QATM method and other state-of-the-art methods. These results confirm the use of QATM, which gives 1-to-1, 1-to-many, and many-to-many matching cases different weights, not only finds more accurate matched regions in the search image, but also reduces the responses in unmatched cases. For example, in the last row, when a nearly homogeneous negative template is given, the proposed QATM method is the only one that tends to give low scores, while others still returns high responses. 

Finally, the matching speed also matters. We thus estimate the processing speed (sec/sample) for each method using the entire OTB dataset. All evaluations are based on an {\tt Intel(R) Xeon(R) E5-4627 v2} CPU and a {\tt GeForce GTX 1080 Ti} GPU respectively. Table~\ref{tab:time} compares the estimated time complexity of different methods. Though QATM contains relative expensive \texttt{softmax} operation, its DNN compatible nature makes GPU processing feasible, which clearly is the fastest method.

\begin{table}[!h]
    \centering
    \small
    \setlength\tabcolsep{2pt}
    \begin{tabular}{c|ccccc|cc}\hline
     \textbf{Methods} & \textbf{SSD} & \textbf{NCC} & \textbf{BBS} & \textbf{DDIS} & \textbf{CoTM}& \multicolumn{2}{c}{\textbf{QATM}} \\\hline
     Backend & \multicolumn{5}{c|}{CPU}     & CPU & GPU\\
         Average (sec.) &  1.1 & 1.5 & 15.3 & 2.6 & 47.7 & 27.4 & {\bf 0.3}\\
         StandDev (sec.) & 0.47 & 0.53 & 13.10 & 2.29 & 18.50 & 17.80 & {\bf 0.12}\\
         \hline
    \end{tabular}
    \caption{Time complexity comparisons. (Time for feature extraction is excluded)}
    \label{tab:time}
\end{table}

\section{Learable QATM Performance}\label{sec:qtam-deep}
In this section, we focus on use the proposed QATM as a differentiable layer with learnable parameters in different template matching applications. 

%=============================================================================================
\subsection{QATM for Image-to-GPS Verification}
%=============================================================================================
The image-to-GPS verification (IGV) task attempts to verify whether a given image is taken as the claimed GPS location through visual verification. IGV first uses the claimed location to find a reference panorama image in a third-party database, \eg \textit{Google StreetView}, and then take both the given image and the reference as network inputs to verify visual contents via template matching and produces the verification decision. The major challenges of the IGV task compared to the classic template matching problem are (1) only a small unknown portion visual content in the query image can be verified in the reference image, and (2) the reference image is a panorama, where the potential matching ROI might be distorted. 

\subsubsection{Baseline and QATM Settings}
\begin{figure}[!h]
    \centering
    \includegraphics[width=.95\linewidth]{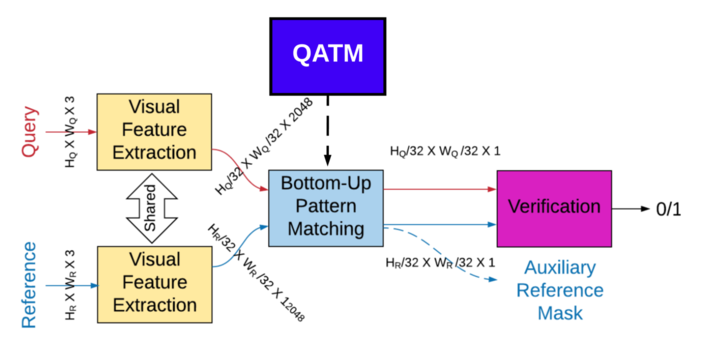}
    \caption{The baseline network architecture from~\cite{cheng2018ACCV}, and the QATM version. The dashed arrows indicate the replacement relationship.}
    \label{fig:gps}
\end{figure}

To understand the QATM performance in the IGV task, we use the baseline method \cite{cheng2018ACCV}
% THIS LINK IS NOT PUBLIC AND MIGHT BE A VIOLATION OF PEER REVIEW RULES. Will COMMENT FOR NOW:
%\footnote{\textcolor{red}{https://gitlab.vista.isi.edu/chengjia/image-GPS}}
, and repeat its network training, data augmentation, evaluation \etc, except that we replace its \textit{Bottom-up Pattern Matching} module with the proposed \texttt{NeuralNetQATM} layer (blue box in Fig.~\ref{fig:gps}). 

% QATM replace BUPM

The \textit{Bottom-up Pattern Matching} module first computes the cosine similarity between two image features, and then pools out the maximum response only. More precisely, its matching score for a patch $s$ given the template \img{T} relies on Eq.~\eqref{eq:bump},
\begin{equation}\label{eq:bump}
    R(s|\mathbf{T}) = \max\{\rho(f_t,f_s) | t \in \mathbf{T}\}
\end{equation}
while the QATM version relies on Eq.~\eqref{eq:qs}.

\subsubsection{Performance Comparison}
To evaluate QATM performance, we reuse the two dataset used by \cite{cheng2018ACCV}, namely the Shibuya and Wikimedia Common dataset, both of which contain balanced positive and negative samples. Comparison results are listed in Table~\ref{tab:gps}. 
\begin{table}[!h]
    \centering
    \small
    \begin{tabular}{c|c|c}
    \hline
        & \textbf{Wikimedia Common}  & \textbf{Shibuya} \\ \hline
       %  &  ROC AUC & Avg. Precision & ROC AUC & Avg. Precision\\
        NetVLAD~\cite{arandjelovic2016netvlad} & 0.819 / 0.847 & 0.634 / 0.638  \\
        DELF~\cite{noh2017largescale} & 0.800 / 0.802 & 0.607 / 0.621 \\
        PlacesCNN~\cite{zhou2017places} & 0.656 / 0.654 & 0.592 / 0.592 \\
        BUPM$^*$~\cite{cheng2018ACCV} & \textbf{0.864} / \textbf{0.886} & 0.764 / 0.781 \\\hline
        QATM & 0.857 / \textbf{0.886} & \textbf{0.777} / \textbf{0.801} \\
    \hline
    \end{tabular}
    \caption{Image-to-GPS verification performance comparisons. Performance scores are reported in the (ROC-AUC / Avg. precision) format. ($^*$ indicates the baseline network.) }
    \label{tab:gps}
\end{table}
The proposed QATM solution outperforms the baseline BUMP method on the more difficult Shibuya dataset, while slightly worse on the Wikimedia Common dataset. This is likely attributed to the fact that the \textit{Verification} (see Fig.~\ref{fig:gps}) in the baseline method is proposed to optimize the BUMP performance but not the QATM performance, and thus the advantage of using QATM has not fully transfer to the verification task.

\begin{figure}[!h]
    \centering\scriptsize
    \includegraphics[width=.995\linewidth]{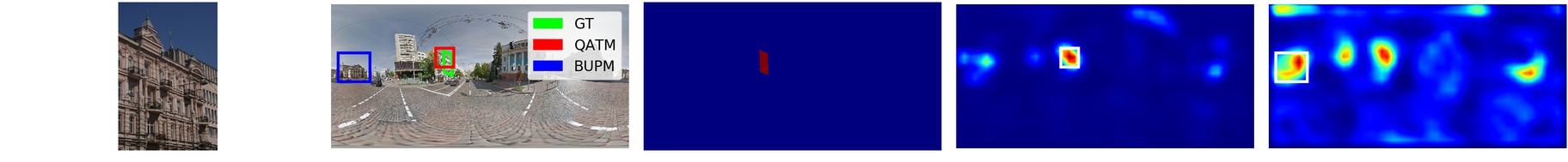}\\
    \includegraphics[width=.995\linewidth]{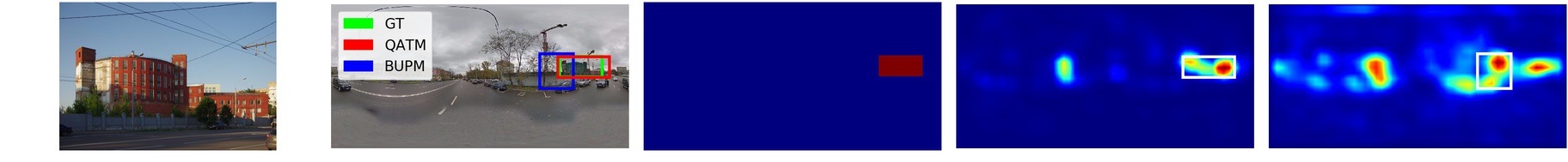}\\
    \includegraphics[width=.995\linewidth]{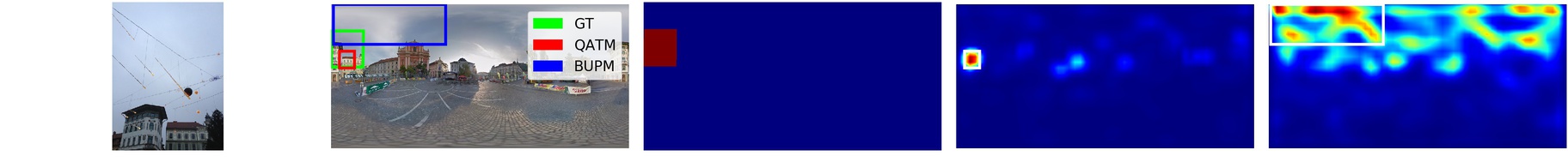}\\
    \includegraphics[width=.995\linewidth]{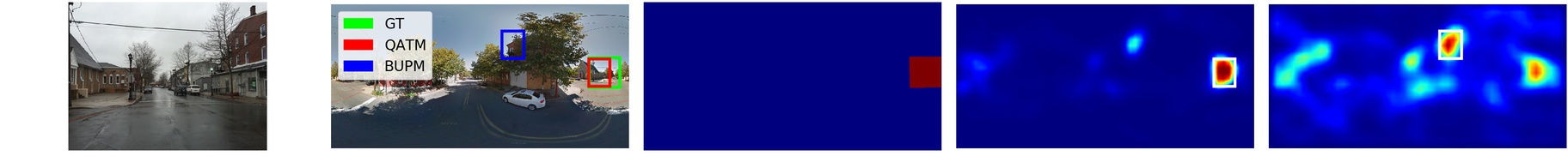}\\
    \includegraphics[width=.995\linewidth]{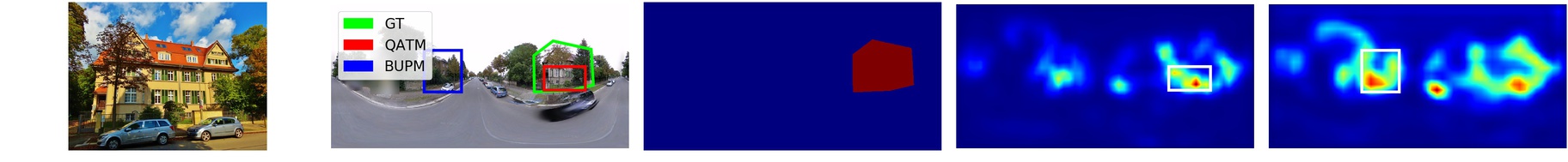}\\
    \includegraphics[width=.995\linewidth]{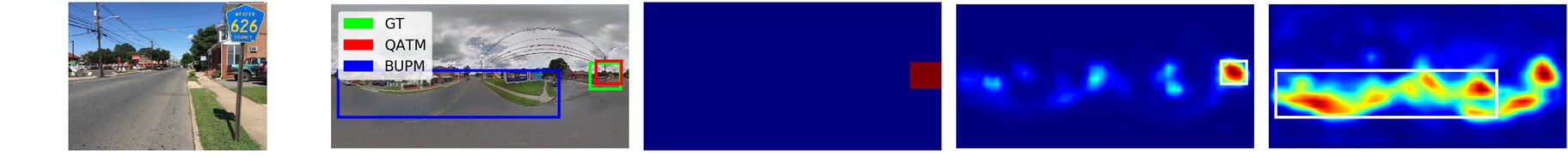}\\
    \includegraphics[width=.995\linewidth]{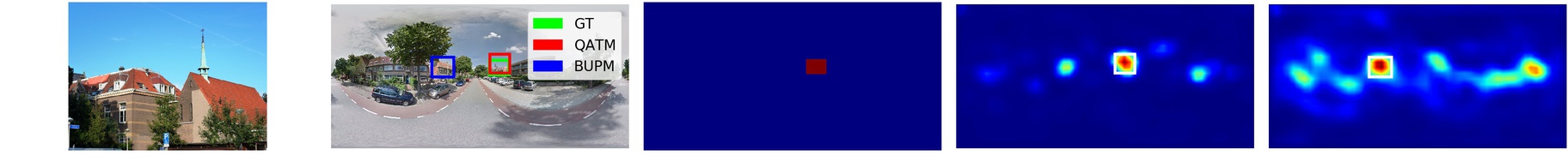}\\
    \includegraphics[width=.995\linewidth]{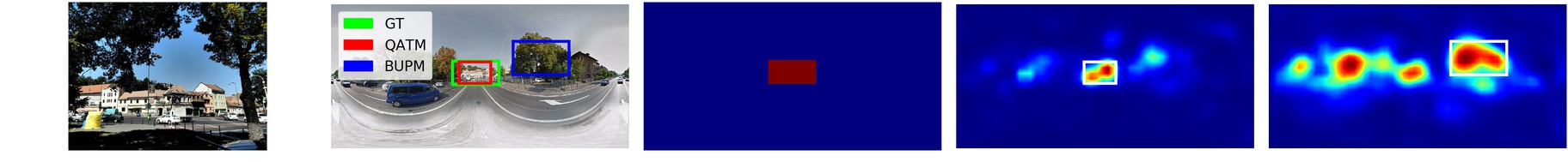}\\
    \includegraphics[width=.995\linewidth]{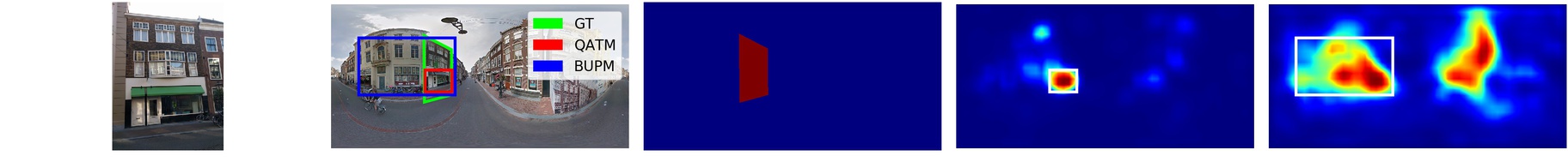}\\
    \includegraphics[width=.995\linewidth]{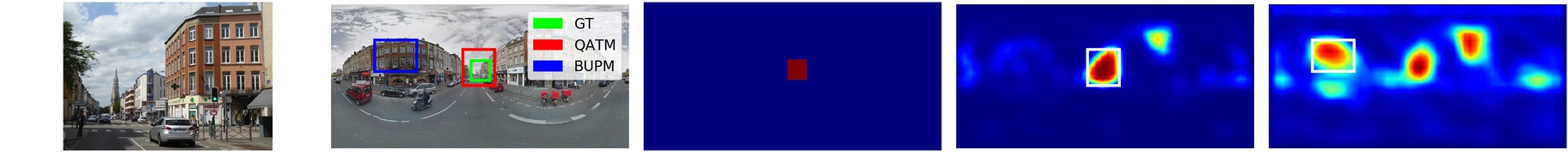}\\
    \includegraphics[width=.995\linewidth]{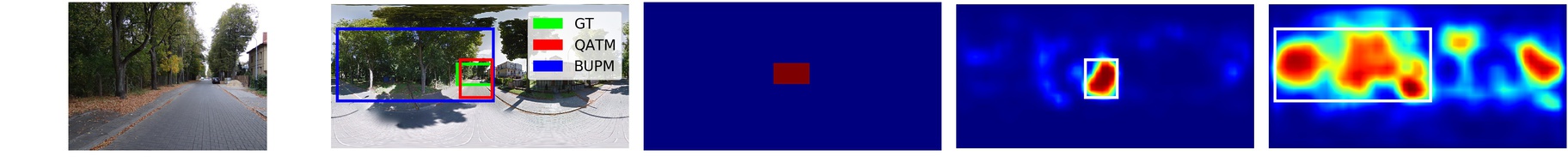}\\
    \includegraphics[width=.995\linewidth]{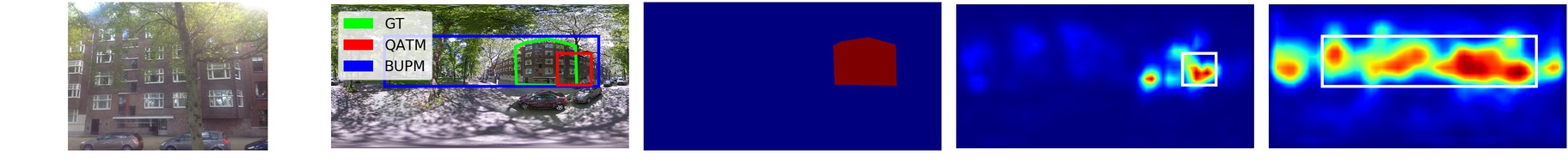}\\
    \includegraphics[width=.995\linewidth]{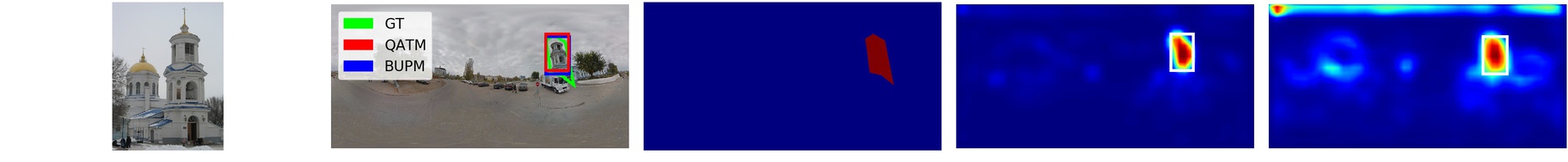}\\
    \includegraphics[width=.995\linewidth]{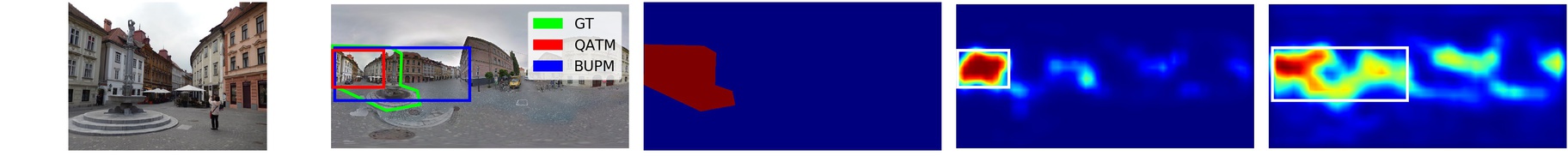}\\
    %\begin{tabular}{@{}p{.2\linewidth}p{.2\linewidth}p{.2\linewidth}p{.2\linewidth}p{.2\linewidth}@{}}
    \begin{tabular*}{\linewidth}{p{.1cm}@{\extracolsep{\fill}}ccccc}
       &Query&
       Search Pano. &
       \textcolor{green}{GT} &
       \textcolor{red}{QATM} \; &
       \textcolor{blue}{BUPM} \quad \;
    \end{tabular*}
    \caption{Qualitative image-to-GPS results. Columns from left to right are: the query image, the reference panorama image with predicted bounding boxes overlaid (\textcolor{green}{GT}, the proposed \textcolor{red}{QATM}, and the baseline \textcolor{blue}{BUPM}), and the response maps of ground truth mask, QATM-improved, and baseline, respectively.}
    \label{fig:gpsResults}
\end{figure}

We therefore annotate the matched regions in terms of polygon bounding boxes for the Wikimedia Common dataset for better evaluating the matching performance. These annotations will be released. With the help of these ground truth masks, we are able to fairly compare the proposed QATM and BUMP only on the localization task, which is to predict the matched region in a panorama image. These results are shown in Table~\ref{tab:gps2}, and the QATM improves the BUMP localization performance by 21\% relatively for both $F_1$ and IoU measure, respectively. The superiority of QATM for localization can be further confirmed in qualitative results shown in Fig.~\ref{fig:gpsResults}, where the QATM-improved version produces much cleaner response maps than the baseline BUMP method.

\begin{table}[!htbp]
\small
    \centering
    \begin{tabular}{c|cc}
    \hline
       \textbf{Wikimedia Common} & \textbf{F1} & \textbf{IoU} \\ \hline
        BUPM  & 0.33 & 0.24\\
        \hline
        QATM  & \textbf{0.40} & \textbf{0.29} \\
    \hline
    \end{tabular}
    \caption{Localization performance comparisons. Performance scores are averaged over the entire dataset.}
    \label{tab:gps2}
\end{table}

%=============================================================================================
\subsection{QATM for  Semantic Image Alignment}
%=============================================================================================
The overall goal for the semantic image alignment (SIA) task is to wrap a given image such that after wrapping it is aligned to a reference image in terms of category-level correspondence. A typical DNN solution for  semantic image alignment task takes two input images, one for wrapping and the other for reference, and commonly output a set of parameters for image wrapping. More detailed descriptions about the problem can be found in \cite{rocco2017convolutional,rocco2018end,Han_2017_ICCV}.\footnote{https://www.di.ens.fr/willow/research/cnngeometric/}\footnote{https://www.di.ens.fr/willow/research/weakalign/}\footnote{https://www.di.ens.fr/willow/research/scnet/}

\subsubsection{Baseline and QATM Settings}
\begin{figure}[!h]
    \centering
    \includegraphics[width=1\linewidth]{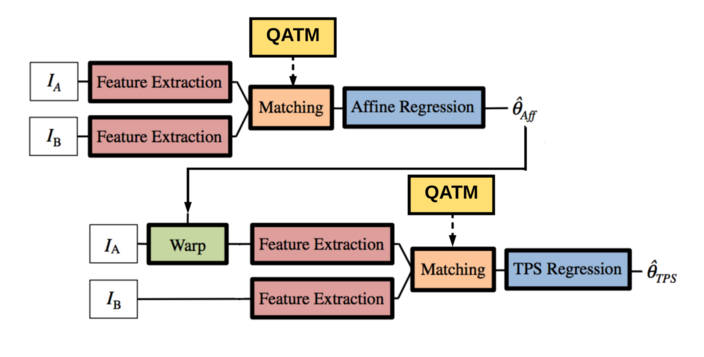}
    \caption{The baseline network architecture from~\cite{rocco2017convolutional}, and the QATM version. The dashed arrows indicate the replacement relationship.}
    \label{fig:isa}
\end{figure}

To understand the QATM performance in the SIA task, we select the baseline method GeoCNN~\cite{rocco2017convolutional}, and mimic all the network related settings, including to network architecture, training dataset, loss function, learning rates, \etc, except that we replace the method's \textit{matching} module (orange box in Fig.~\ref{fig:isa}) with the \texttt{NeuralNetQATM} layer (yellow box in  Fig.~\ref{fig:isa}). 

Unlike in template matching, the SIA task relies on the raw matching scores between all template and search image patches (such that geometric information is implicitly preserved) to regress the wrapping parameters. The \textit{matching} module in ~\cite{rocco2017convolutional} is simply computed as the cosine similarity between two patches, \ie $\rho(\mathbf{S},\mathbf{T})$ (see $\rho_{st}$ in line 4 of Alg.~\ref{alg:qatm}) and use this tensor as the input for regression. As a result, instead of the matching quality maps, we also make the corresponding change that let the proposed \texttt{NeuralNetQATM} produce the raw QATM matching scores, \ie QATM$(\mathbf{S},\mathbf{T})$(see {\texttt{QATM}} in line 8 of Alg.~\ref{alg:qatm}). 

\subsubsection{Performance Comparisons}
\begin{figure}[!h]
    \centering
    \includegraphics[width=.995\linewidth]{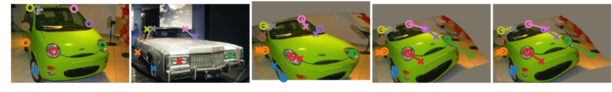}\\
    \includegraphics[width=.995\linewidth]{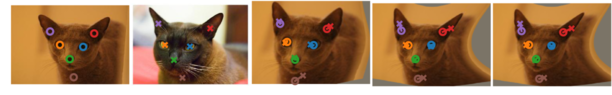}\\
    \includegraphics[width=.995\linewidth]{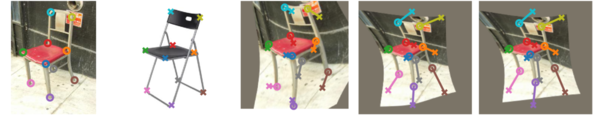}\\
    \includegraphics[width=.995\linewidth]{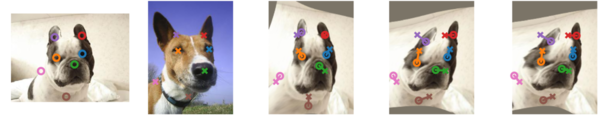}\\
    \includegraphics[width=.995\linewidth]{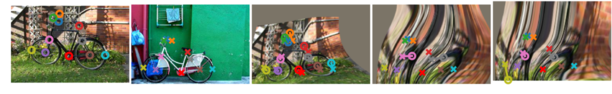}\\
    \includegraphics[width=.995\linewidth]{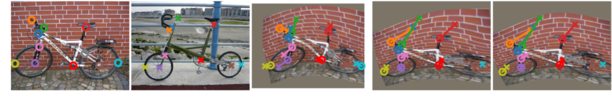}\\
    \includegraphics[width=.995\linewidth]{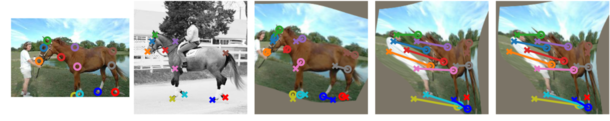}\\
    \includegraphics[width=.995\linewidth]{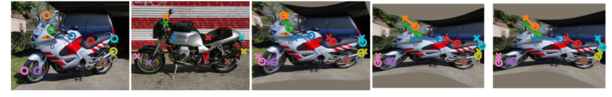}\\
    \includegraphics[width=.995\linewidth]{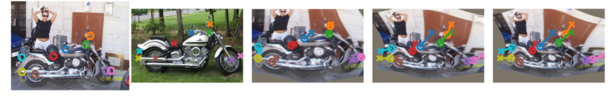}\\
    \includegraphics[width=.995\linewidth]{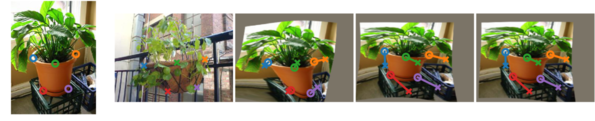}\\
    \includegraphics[width=.995\linewidth]{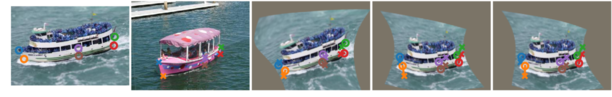}\\
    \includegraphics[width=.995\linewidth]{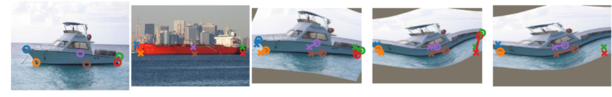}\\
    \includegraphics[width=.995\linewidth]{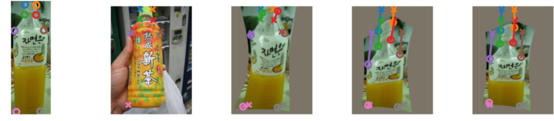}\\
    \includegraphics[width=.995\linewidth]{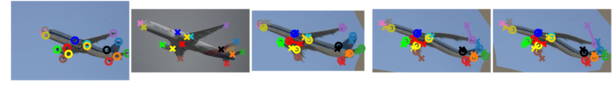}\\

    \caption{Qualitative results on PF-PASCAL dataset. Columns from left to right represent source image, target image, transform results of QATM, GoeCNN\cite{rocco2017convolutional} and \cite{rocco2018end}. Circles and crosses indicate key points on source images and target images.}
    \label{fig:isaResult}
\end{figure}

To fairly compare SIA performance, we follow the evaluation protocols proposed in \cite{Han_2017_ICCV}, which splits the standard PF-PASCAL benchmark into training, validation, and testing subsets with 700, 300, and 300 samples, respectively. The system performance is reported in terms of the percentage of correct key points (PCK)~\cite{yang2013articulated,Han_2017_ICCV}, which counts the percentage of key points whose distance to ground truth is under a threshold after being transformed. The threshold is set to $\tau=0.1$ of image size in the experiment.
Table~\ref{tb.semantic_alignment} compares different methods on this dataset. The proposed QATM method clearly outperforms all baseline methods, and also is the top-ranking method for 7 out of 20 sub-classes. Furthermore,  the SCNet~\cite{Han_2017_ICCV} uses much more advanced features and matching mechanisms than our baseline GeoCNN method. And \cite{rocco2018end} used training subset of PF-PASCAL to fine-tune on GeoCNN with a very small learning rate. However, our results confirm that simply replacing the raw matching scores with those quality-aware scores could lead an larger gain than using more a complicated network without fine-tuning on PF-PASCAL subset. A concurrent work~\cite{rocco2018neighbourhood} adopted a similar idea to re-rank matching score through softmax function as QATM. They reassigned matching score by finding soft mutual nearest neighbour and outperformed QATM when trained on PF-PASCAL subset. More qualitative results can be found in Fig.~\ref{fig:isaResult}

\begin{table}[!h]
    \centering\setlength{\tabcolsep}{2pt}
    \small
    \begin{tabular}{r|ccccc|c}\hline
\textbf{Class}& UCN &     SCNet &  GeoCNN$^*$ &  WSup & NC-Net &QATM  \\
&~\cite{choy2016universal} & ~\cite{Han_2017_ICCV} & ~\cite{rocco2017convolutional} & ~\cite{rocco2018end} &~\cite{rocco2018neighbourhood} & \\ \hline
  \textbf{plane} & 64.8 & {\bf 85.5} & 82.4 &\underline{83.7} & - &83.5  \\
  \textbf{bike} & 58.7 & 84.4 & 80.9 & {\bf 88.0} &- & \underline{86.2}  \\
 \textbf{bird} & 42.8 & 66.3 & {\bf 85.9} & \underline{83.4} &- & 80.7  \\
 \textbf{boat} & 59.6 & \underline{70.8} & 47.2 & 58.3 &- & {\bf 72.2}  \\
 \textbf{bottle} & 47.0 & 57.4 & 57.8 & \underline{68.8} &- & {\bf 78.1}  \\
 \textbf{bus} & 42.2 & 82.7 & 83.1 & {\bf 90.3} &- & \underline{87.4}  \\
 \textbf{car} & 61.0& 82.3 & {\bf 92.8} & \underline{92.3} &- & 91.8  \\
 \textbf{cat} &  45.6 & 71.6 & {\bf 86.9} & 83.7 &- & {\bf 86.9}  \\
 \textbf{chair} & \underline{49.9}& {\bf 54.3} & 43.8 & 47.4 &- & 48.8  \\
 \textbf{cow} &  52.0 & {\bf 95.8} & \underline{91.7} & \underline{91.7} &- & 87.5  \\
 \textbf{d.table} & \underline{48.5} & {\bf 55.2} &28.1 & 28.1 &- & 26.6  \\
 \textbf{dog} & 49.5 & 59.5 & \underline{76.4} & 76.3 &- & {\bf 78.7}  \\
 \textbf{horse} & 53.2 & 68.6 & 70.2 & \underline{77.0} &- & {\bf 77.9}  \\
 \textbf{m.bike} & 72.7 & 75.0 & \underline{76.6} & \underline{76.0} &- & {\bf 79.9}  \\
 \textbf{person} & 53.0 & 56.3 & 68.9 & {\bf 71.4} &- & \underline{69.5}  \\
 \textbf{plant} & 41.4 & 60.4 & 65.7 & {\bf 76.2} &- & \underline{73.3}  \\
 \textbf{sheep} & 83.3 & 60.0 & {\bf 80.0} & {\bf 80.0} &- & {\bf 80.0}  \\
 \textbf{sofa} & 49.0 & {\bf 73.7} & 50.1 & \underline{59.5} &- & 51.6  \\
 \textbf{train} & {\bf 73.0} & \underline{66.5} & 46.3 & 62.3 &- & 59.3  \\
 \textbf{tv} & \underline{66.0} & {\bf 76.7} & 60.6 & 63.9 &- & 64.4  \\\hline
 \textbf{Average}& 55.6 & 72.2 & 71.9 & 75.8 & {\bf 78.9} & \underline{75.9} \\ \hline
    \end{tabular}
    \caption{Semantic image alignment performance comparison on PF-PASCAL. ($^*$ indicates the baseline network.) }
    \label{tb.semantic_alignment}
\end{table}

%=========================================================================================================
%=========================================================================================================

\section{Conclusion}\label{sec:conclusion}
We introduced a novel quality-aware template matching method, QTAM. QTAM is inspired by the fact of natural quality differences among different matching cases. It is also designed in such a way that its matching score accurately reflects the relative matching distinctiveness of the current matching pair against others. More importantly, QTAM is differentiable with a learable paramters, and can easily be implemented with existing common deep learning layers. QTAM can be directly embedded into a DNN model to fulfill the template matching goal.

Our extensive experiments show that when used alone, it outperforms the state-of-the-art template matching methods and produces more accurate matching performance, fewer false alarms, and at least 10x speedup with the help of a GPU. When plugged into existing DNN solutions for template matching related tasks, we demonstrated that it could noticeably improve the scores in both the image semantic alignment tasks, and the image-to-GPS verification task.

\renewcommand{\baselinestretch}{1.0}
\noindent {\small {\bf Acknowledgement}  This work is based on research sponsored by the Defense Advanced Research Projects Agency under agreement number FA8750-16-2-0204. The U.S. Government is authorized to reproduce and distribute reprints for governmental purposes notwithstanding any copyright notation thereon. The views and conclusions contained herein are those of the authors and should not be interpreted as necessarily representing the official policies or endorsements, either expressed or implied, of the Defense Advanced Research Projects Agency or the U.S. Government.\par}
\clearpage

{\small
\bibliographystyle{ieee_fullname}
\bibliography{egbib}

\begin{thebibliography}{10}\itemsep=-1pt

\bibitem{adam2006robust}
Amit Adam, Ehud Rivlin, and Ilan Shimshoni.
\newblock Robust fragments-based tracking using the integral histogram.
\newblock In {\em Proceedings of the IEEE Conference on Computer Vision and
  Pattern Recognition}, volume~1, pages 798--805. IEEE, 2006.

\bibitem{alaei2014complete}
Alireza Alaei and Mathieu Delalandre.
\newblock A complete logo detection/recognition system for document images.
\newblock In {\em Proceedings of International Workshop on Document Analysis
  Systems)}, pages 324--328. IEEE, 2014.

\bibitem{arandjelovic2016netvlad}
Relja Arandjelovic, Petr Gronat, Akihiko Torii, Tomas Pajdla, and Josef Sivic.
\newblock Netvlad: Cnn architecture for weakly supervised place recognition.
\newblock In {\em Proceedings of the IEEE Conference on Computer Vision and
  Pattern Recognition}, pages 5297--5307, 2016.

\bibitem{bai2016exploiting}
Min Bai, Wenjie Luo, Kaustav Kundu, and Raquel Urtasun.
\newblock Exploiting semantic information and deep matching for optical flow.
\newblock In {\em European Conference on Computer Vision}, pages 154--170.
  Springer, 2016.

\bibitem{boia2016logo}
Raluca Boia, Corneliu Florea, Laura Florea, and Radu Dogaru.
\newblock Logo localization and recognition in natural images using homographic
  class graphs.
\newblock {\em Machine Vision and Applications}, 27(2):287--301, 2016.

\bibitem{brown2007automatic}
Matthew Brown and David~G Lowe.
\newblock Automatic panoramic image stitching using invariant features.
\newblock {\em International Journal of Computer Vision}, 74(1):59--73, 2007.

\bibitem{cheng2018ACCV}
Jiaxing Cheng, Yue Wu, Wael AbdAlmageed, and Prem Natarajan.
\newblock Image-to-gps verification through a bottom-up pattern matching
  network.
\newblock In {\em Asian Conference on Computer Vision}. Springer, 2018.

\bibitem{choy2016universal}
Christopher~B Choy, JunYoung Gwak, Silvio Savarese, and Manmohan Chandraker.
\newblock Universal correspondence network.
\newblock In {\em Advances in Neural Information Processing Systems}, pages
  2414--2422, 2016.

\bibitem{comaniciu2003kernel}
Dorin Comaniciu, Visvanathan Ramesh, and Peter Meer.
\newblock Kernel-based object tracking.
\newblock {\em IEEE Transactions on Pattern Analysis and Machine Intelligence},
  25(5):564--577, 2003.

\bibitem{coughlan2000efficient}
James Coughlan, Alan Yuille, Camper English, and Dan Snow.
\newblock Efficient deformable template detection and localization without user
  initialization.
\newblock {\em Computer Vision and Image Understanding}, 78(3):303--319, 2000.

\bibitem{dekel2015best}
Tali Dekel, Shaul Oron, Michael Rubinstein, Shai Avidan, and William~T Freeman.
\newblock Best-buddies similarity for robust template matching.
\newblock In {\em Proceedings of the IEEE Conference on Computer Vision and
  Pattern Recognition}, pages 2021--2029, 2015.

\bibitem{felzenszwalb2010object}
Pedro~F Felzenszwalb, Ross~B Girshick, David McAllester, and Deva Ramanan.
\newblock Object detection with discriminatively trained part-based models.
\newblock {\em IEEE transactions on Pattern Analysis and Machine Intelligence},
  32(9):1627--1645, 2010.

\bibitem{Han_2017_ICCV}
Kai Han, Rafael~S. Rezende, Bumsub Ham, Kwan-Yee~K. Wong, Minsu Cho, Cordelia
  Schmid, and Jean Ponce.
\newblock Scnet: Learning semantic correspondence.
\newblock In {\em Proceedings of the IEEE International Conference on Computer
  Vision}, Oct 2017.

\bibitem{kat2018matching}
Rotal Kat, Roy Jevnisek, and Shai Avidan.
\newblock Matching pixels using co-occurrence statistics.
\newblock In {\em Proceedings of the IEEE Conference on Computer Vision and
  Pattern Recognition}, pages 1751--1759, 2018.

\bibitem{luo2016efficient}
Wenjie Luo, Alexander~G Schwing, and Raquel Urtasun.
\newblock Efficient deep learning for stereo matching.
\newblock In {\em Proceedings of the IEEE Conference on Computer Vision and
  Pattern Recognition}, pages 5695--5703, 2016.

\bibitem{malti2013monocular}
Abed Malti, Richard Hartley, Adrien Bartoli, and Jae-Hak Kim.
\newblock Monocular template-based 3d reconstruction of extensible surfaces
  with local linear elasticity.
\newblock In {\em Proceedings of the IEEE Conference on Computer Vision and
  Pattern Recognition}, pages 1522--1529, 2013.

\bibitem{noh2017largescale}
Hyeonwoo Noh, Andre Araujo, Jack Sim, Tobias Weyand, and Bohyung Han.
\newblock Largescale image retrieval with attentive deep local features.
\newblock In {\em Proceedings of the IEEE International Conference on Computer
  Vision}, pages 3456--3465, 2017.

\bibitem{rocco2017convolutional}
Ignacio Rocco, Relja Arandjelovic, and Josef Sivic.
\newblock Convolutional neural network architecture for geometric matching.
\newblock In {\em Proceedings of the IEEE Conference on Computer Vision and
  Pattern Recognition}, volume~2, 2017.

\bibitem{rocco2018end}
Ignacio Rocco, Relja Arandjelovic, and Josef Sivic.
\newblock End-to-end weakly-supervised semantic alignment.
\newblock In {\em Proceedings of the IEEE Conference on Computer Vision and
  Pattern Recognition}, pages 6917--6925, 2018.

\bibitem{rocco2018neighbourhood}
Ignacio Rocco, Mircea Cimpoi, Relja Arandjelovi{\'c}, Akihiko Torii, Tomas
  Pajdla, and Josef Sivic.
\newblock Neighbourhood consensus networks.
\newblock In {\em Advances in Neural Information Processing Systems}, pages
  1658--1669, 2018.

\bibitem{ryan2015examination}
Michael Ryan and Novita Hanafiah.
\newblock An examination of character recognition on id card using template
  matching approach.
\newblock {\em Procedia Computer Science}, 59:520--529, 2015.

\bibitem{scharstein2002taxonomy}
Daniel Scharstein and Richard Szeliski.
\newblock A taxonomy and evaluation of dense two-frame stereo correspondence
  algorithms.
\newblock {\em International Journal of Computer Vision}, 47(1-3):7--42, 2002.

\bibitem{seitz2006comparison}
Steven~M Seitz, Brian Curless, James Diebel, Daniel Scharstein, and Richard
  Szeliski.
\newblock A comparison and evaluation of multi-view stereo reconstruction
  algorithms.
\newblock In {\em Proceedings of the IEEE Conference on Computer Vision and
  Pattern Recognition}, volume~1, pages 519--528. IEEE, 2006.

\bibitem{simakov2008summarizing}
Denis Simakov, Yaron Caspi, Eli Shechtman, and Michal Irani.
\newblock Summarizing visual data using bidirectional similarity.
\newblock In {\em Proceedings of the IEEE Conference on Computer Vision and
  Pattern Recognition}, pages 1--8. IEEE, 2008.

\bibitem{szeliski2007image}
Richard Szeliski et~al.
\newblock Image alignment and stitching: A tutorial.
\newblock {\em Foundations and Trends in Computer Graphics and Vision},
  2(1):1--104, 2007.

\bibitem{talmi2017template}
Itamar Talmi, Roey Mechrez, and Lihi Zelnik-Manor.
\newblock Template matching with deformable diversity similarity.
\newblock In {\em Proceedings of the IEEE Conference on Computer Vision and
  Pattern Recognition}, pages 1311--1319, 2017.

\bibitem{tang2016multi}
Siyu Tang, Bjoern Andres, Mykhaylo Andriluka, and Bernt Schiele.
\newblock Multi-person tracking by multicut and deep matching.
\newblock In {\em European Conference on Computer Vision}, pages 100--111.
  Springer, 2016.

\bibitem{thewlis2016fully}
James Thewlis, Shuai Zheng, Philip~HS Torr, and Andrea Vedaldi.
\newblock Fully-trainable deep matching.
\newblock In {\em British Machine Vision Conference}, 2016.

\bibitem{trier1996feature}
Oivind~Due Trier, Anil~K Jain, Torfinn Taxt, et~al.
\newblock Feature extraction methods for character recognition-a survey.
\newblock {\em Pattern Recognition}, 29(4):641--662, 1996.

\bibitem{wu2017deep}
Yue Wu, Wael Abd-Almageed, and Prem Natarajan.
\newblock Deep matching and validation network: An end-to-end solution to
  constrained image splicing localization and detection.
\newblock In {\em Proceedings of the ACM on Multimedia Conference}, pages
  1480--1502. ACM, 2017.

\bibitem{BusterNet}
Yue Wu, Wael Abd-Almageed, and Prem Natarajan.
\newblock Busternet: Detecting copy-move image forgery with source/target
  localization.
\newblock In {\em European Conference on Computer Vision}, September 2018.

\bibitem{wu2013online}
Yi Wu, Jongwoo Lim, and Ming-Hsuan Yang.
\newblock Online object tracking: A benchmark.
\newblock In {\em Proceedings of the IEEE Conference on Computer Vision and
  Pattern Recognition}, pages 2411--2418, 2013.

\bibitem{yang2013articulated}
Yi Yang and Deva Ramanan.
\newblock Articulated human detection with flexible mixtures of parts.
\newblock {\em IEEE transactions on Pattern Analysis and Machine Intelligence},
  35(12):2878--2890, 2013.

\bibitem{yuille1992feature}
Alan~L Yuille, Peter~W Hallinan, and David~S Cohen.
\newblock Feature extraction from faces using deformable templates.
\newblock {\em International Journal of Computer Vision}, 8(2):99--111, 1992.

\bibitem{zhang2014partial}
Tianzhu Zhang, Kui Jia, Changsheng Xu, Yi Ma, and Narendra Ahuja.
\newblock Partial occlusion handling for visual tracking via robust part
  matching.
\newblock In {\em Proceedings of the IEEE Conference on Computer Vision and
  Pattern Recognition}, pages 1258--1265, 2014.

\bibitem{zhang2015robust}
Tianzhu Zhang, Si Liu, Narendra Ahuja, Ming-Hsuan Yang, and Bernard Ghanem.
\newblock Robust visual tracking via consistent low-rank sparse learning.
\newblock {\em International Journal of Computer Vision}, 111(2):171--190,
  2015.

\bibitem{zhang2017learning}
Xu Zhang, X~Yu Felix, Sanjiv Kumar, and Shih-Fu Chang.
\newblock Learning spread-out local feature descriptors.
\newblock In {\em Proceedings of the IEEE International Conference on Computer
  Vision}, pages 4605--4613, 2017.

\bibitem{zhang2018heated}
Xu Zhang, Felix~Xinnan Yu, Svebor Karaman, Wei Zhang, and Shih-Fu Chang.
\newblock Heated-up softmax embedding.
\newblock {\em arXiv preprint arXiv:1809.04157}, 2018.

\bibitem{zhou2017places}
Bolei Zhou, Agata Lapedriza, Aditya Khosla, Aude Oliva, and Antonio Torralba.
\newblock Places: A 10 million image database for scene recognition.
\newblock {\em IEEE Transactions on Pattern Analysis and Machine Intelligence},
  2017.

\end{thebibliography}
}

\end{document}